\documentclass[10pt, a4paper]{article}
\usepackage{lrec}
\usepackage{multibib}
\newcites{languageresource}{Language Resources}
\usepackage{graphicx}
\usepackage{tabularx}
\usepackage{soul}
\usepackage{algorithm}
\usepackage{algorithmic}
\usepackage{epstopdf}
\usepackage[latin1]{inputenc}

\usepackage{hyperref}
\usepackage{xstring}

\title{Dynamic Oracle for Neural Machine Translation in Decoding Phase}

\name{Zi-Yi Dou, Hao Zhou, Shu-Jian Huang, Xin-Yu Dai, Jia-Jun Chen}

\address{National Key Laboratory for Novel Software Technology, Nanjing University \\
         Nanjing, 210023, China\ \\
         141242042@smail.nju.edu.cn, \{zhouh, huangsj, daixy, chenjj\}@nlp.nju.edu.cn\\}

\abstract{
The past several years have witnessed the rapid progress of end-to-end Neural Machine Translation (NMT). 
However, there exists discrepancy between training and inference in NMT when decoding, which may lead to serious problems since the model might be in
a part of the state space it has never seen during training. To address the issue, Scheduled Sampling has been proposed. However, there are certain limitations in Scheduled Sampling and we propose two dynamic oracle-based methods to improve it. We
manage to mitigate the discrepancy by changing the training process towards a less guided scheme and meanwhile aggregating the oracle's demonstrations. Experimental results show that
the proposed approaches improve translation quality over standard NMT system. \\ \newline  }

\begin{document}

\maketitleabstract

\section{Introduction}
Neural networks have been widely used contemporarily and have achieved great performance
on a variety of fields like sentiment analysis \cite{Santos2014Deep} and visual object 
recognition \cite{ciregan2012multi}.
For sequential problems, 
recurrent neural networks can be applied to process sequences.
To address issues like long term dependencies in the data \cite{bengio1994learning}, 
the Long Short-Term Memory (LSTM) \cite{hochreiter1997long} or Gated Recurrent Unit (GRU) can be used to tackle the problem \cite{cho2014learning}.
A straightforward application of the LSTM and GRU architecture have already
shown impressive performance in several difficult tasks, including machine translation \cite{sutskever2014sequence},
and image captioning \cite{vinyals2015show}.

Generally, a basic sequence-to-sequence model consists of two recurrent neural networks:
 an encoder that processes the input and a decoder that generates the output \cite{cho2014learning}.
 In many applications of sequence-to-sequence models, at inference time, the output
 of the decoder at time $t$ is fed back and becomes the input of decoder at time $t+1$.
 However, during training, it is more common to provide the correct input to the decoder 
 at every time-step even if the decoder made a mistake before, which leads to a discrepancy
 between how the model is used at training and inference. 
As has been pointed out by Bengio {\it et. al.} \shortcite{bengio2015scheduled}, 
although this discrepancy can be mitigated by the use of a beam search heuristic maintaining 
several generated target sequences, for continuous state space models like recurrent neural networks,
there is no dynamic programming approach, so the effective number of sequences considered 
remains small. 
The main problem is that mistakes made earlier in the sequence generation process
are fed as input to the model and can be quickly amplified because the model might be
in a part of the state space it has never seen at training time \cite{bengio2015scheduled}.

There are several existing methods to bridge the gap between training and inference. Bengio {\it et. al.} \shortcite{bengio2015scheduled}
propose a method called {\it Scheduled Sampling}.
Since the main difference between training and inference for sequence prediction tasks when
predicting token $y_t$ is whether we use the true previous token $y_{t-1}$ or an estimate $\hat{y}_{t-1}$
coming from the model itself, 
they apply a sampling mechanism that will randomly decide, 
during  training, whether to use the true previous token $y_{t-1}$ or an estimate $\hat{y}_{t-1}$ coming from the
model itself. 
Specifically, for every token, they flip a coin and decide whether to use the true previous token or an estimate coming from the model itself. Their general framework is shown in \textbf{Figure 1}.
\begin{figure}
\centering
\includegraphics[width=7.5cm]{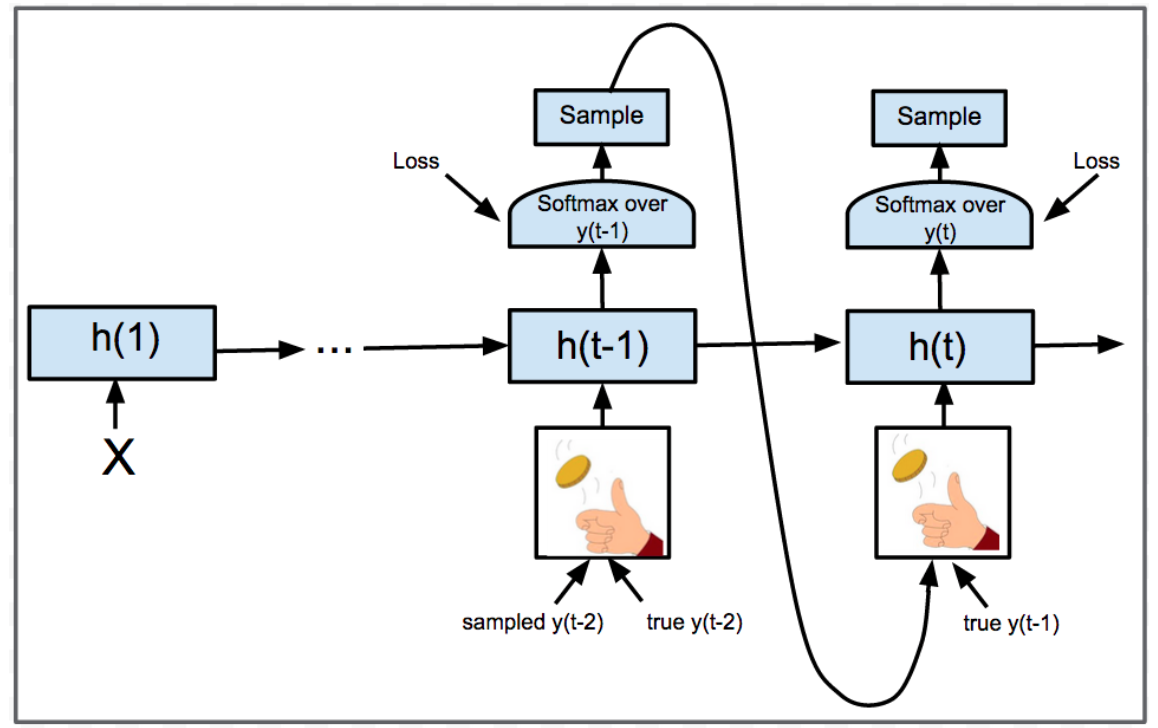}
\caption{Scheduled Samping}  
\end{figure}
At the beginning of training, the model is not well-trained and thus selecting more often the true previous token should be helpful; 
whereas at the end of training the sampling strategy should favor sampling from the model
more often, as this corresponds to the true inference situation. Therefore,
the probability of using $y_{t-1}$ will be initially high and then decrease during training. In their paper, they choose three different functions to model the probability of using $y_{t-1}$ with respect to training time and they achieve competitive results.

Although {\it Scheduled Sampling} has been proved to be helpful on several tasks \cite{bengio2015scheduled}, for machine translation tasks, in our experiments it does not show promising
performance. We have done a few research and found out a dissatisfactory characteristic about {\it Scheduled Sampling}. To illustrate, as we can see from \textbf{Figure 2}, since the reference has been altered, the original correct inputs would not 
be accurate and therefore it is unwise to still provide the model with the original inputs. 

In another words, {\it Scheduled Sampling} still uses the gold word for training whereas it would be no longer correct, which is a static oracle fashion.
We believe that if we can use a certain strategy, such as {\it dynamic oracle}, to always provide the model with 
accurate inputs, the performance will be better.

\begin{figure}
\centering
\includegraphics[width=7.5cm]{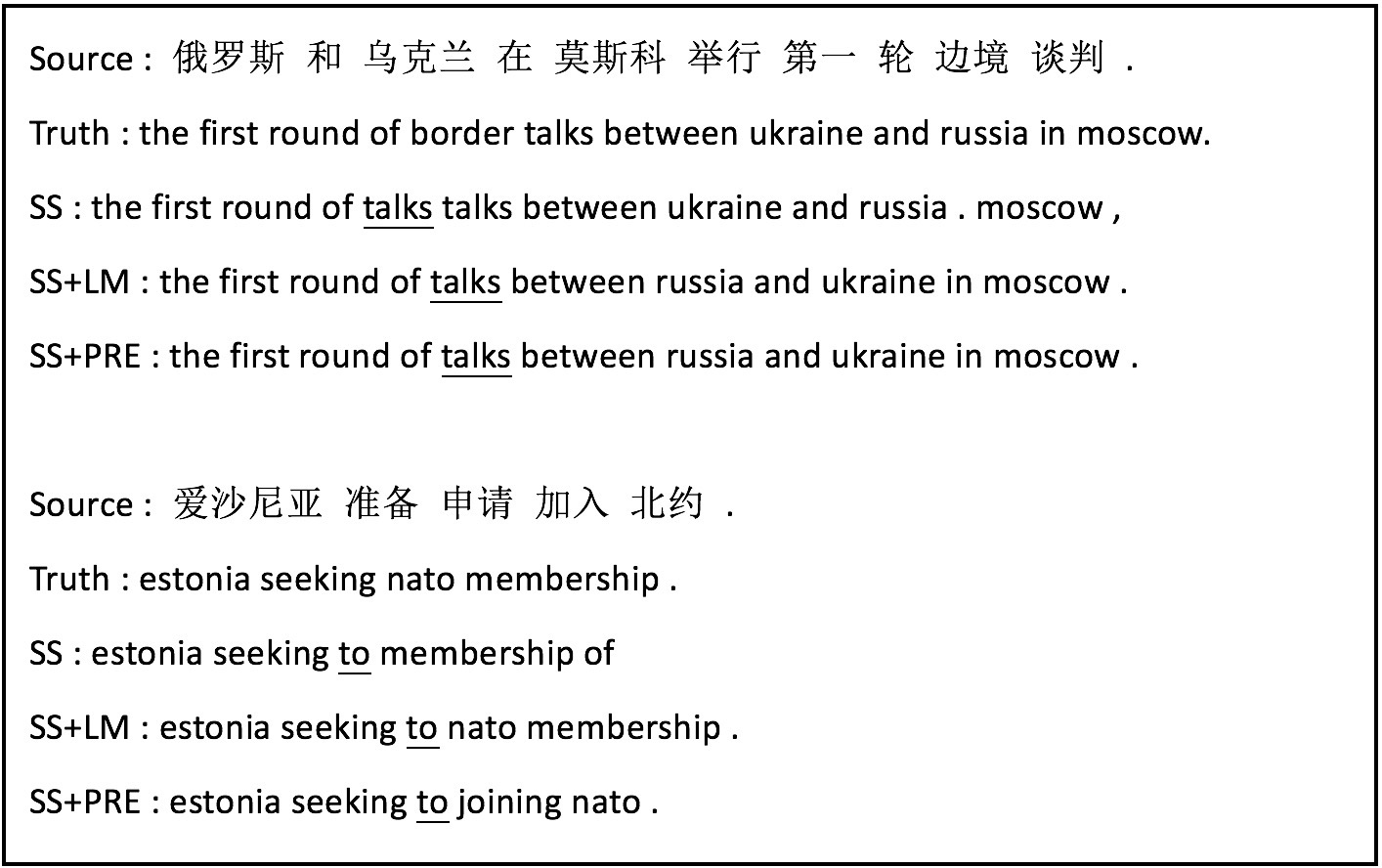}
\caption{\label{Fig:1} Sampling Results}  
\end{figure}

In this paper, based on the aforementioned idea, we propose two methods which use {\it dynamic oracle} to solve the issue mentioned above. To illustrate, once we decide to use an estimated token coming
from the model itself, our {\it oracle} can help us select the next suitable word and feed it into the model. Therefore, 
even during training time the truth have been completely altered, our method will always provide the model with the best suitable token in the following time-steps. 
In this way, we hope that the aforementioned problem in {\it Scheduled Sampling} would be mitigated. 

Here we develop two strategies to implement the {\it dynamic oracle}, one is based on language model and the other is based on pre-trained neural machine translation system. These two methods could manage to feed the correct word into the model so that they may enhance the performance of  {\it Scheduled Sampling}.

To verify the effectiveness of our methods, we conduct experiments on Chinese-English datasets and the experimental results indicate that
our method can achieve $+1.06$ BLEU improvements. 

\section{Related Work}

To the best of our knowledge, Goldberg {\it et. al.}  \shortcite{goldberg2012dynamic}
first define the concept of {\it dynamic oracle} and propose an online
algorithm for parsing problems,
, which provides a set of optimal transitions for every valid parser configuration. 
For configurations
which are not part of a gold derivation, their dynamic oracle permits all
transitions that can lead to a tree with minimum loss compared 
to the gold tree. Based on their approach, several other methods using dynamic oracle
have been proposed \cite{goldberg2013training} \cite{gomez2014polynomial}.
 However, their work in the field of parsing cannot 
be directly applied in neural machine translation.

To mitigate the discrepancy between training and inference, Daume {\it et al.} \shortcite{halsearch} introduce SEARN, 
 which aims to tackle the problems that training
 examples might be different from actual test examples. 
 They show that
 structured prediction can be mapped into a search setting using language
 from reinforcement learning, and known techniques for reinforcement learning can give
 formal performance bounds on the structured prediction task.
 In addition, Dataset Aggregation (DAgger) \cite{Ross2011A} is another method which adds on-policy
 samples to its dataset and then re-optimizes the policy
 by asking human to label these new data.

\section{Proposed Methods}
In this section, we first give a brief introduction of neural machine translation. And then we present the general framework for our algorithms. At last, we describe our two methods respectively, namely 
 language model guided scheduled sampling and pre-trained model guided scheduled sampling.
 
 \subsection{Neural Machine Translation}
 
 Neural machine translation aims to directly model the conditional probability $p(Y|X)$ of translating 
 a source sentence, $x_1, ..., x_n$, to a target sentence, $y_1, ..., y_m$. Generally, it accomplishes this goal through
 an encoder-decoder framework \cite{kalchbrenner2013recurrent}.
Basically, the encoder generates a context vector for each source sentence and then the decoder 
outputs a translation, one target word at a time.

During training when we are decoding, we always provide the model with true previous token at every time step. Mini-batch stochastic gradient
descent is applied to look for a set of 
parameters $\theta^{*}$ that 
maximizes the log likelihood of
producing the correct target sentence.
Specifically, given a batch of training pairs $\{(X^i,Y^i)\}$, 
we aim to find $\theta^{*}$ which satisfies:
\begin{equation}
\theta^* =\arg \max_{\theta} \sum_{(X^i,Y^i)} \log p(Y^i|X^i;\theta)
\end{equation}

Whereas during inference the model can generate the target sentence one token
at a time, advancing time by one step.
This procedure will continue until an $<$EOS$>$ token is generated.
Since at time $t$ we do not have access to the true previous token,
normally we just feed the model with the most likely token given our model at time $t$.
To search for the sentence with the highest probability, beam search is often used.

The log conditional probability can be decomposed as:
\begin{equation}
\log p(Y|X)=\sum_{t=1}^m \log p(y_t | y_{<t} , C),
\end{equation}
where $C$ is the context vector for source sentence.

The log conditional probability $\log p(y_t | y_{<t} , C)$ is computed in different ways according to the choice 
of the context $C$ at time $i$.  Bahdanau {\it et. al.} \shortcite{cho2014learning} use different context $c_i$ at different time step while Cho {\it et. al.}  \shortcite{bahdanau2014neural} choose
$C= \textbf{h}_{T_x}$.

The architecture of recurrent neural network may differ in terms of architecture and type. 
For example, there can be unidirectional, bidirectional or deep multi-layer RNN; and RNN type 
can be LSTM \cite{hochreiter1997long} or the GRU\cite{cho2014learning}.

\subsection{General Framework and Definitions}
The general goal of our work is to mitigate the discrepancy between training and inference when using recurrent neural network.
To achieve this goal, we borrow the idea from {\it Scheduled Sampling}. Specifically, during training when decoding, there is a possibility that we feed our model with an estimate $\hat{y}_{t-1}$ coming from the model itself rather than true previous token $y_{t-1}$. Different from {\it Scheduled Sampling}, once we decide to use $\hat{y}_{t-1}$, the original ground-truth would not be accurate any more and thus we use {\it dynamic oracle} to generate the next suitable token $y^*_{t-1}$ and regard it as our true token. The pseudo-code of our general framework is shown in Algorithm 1.

\begin{algorithm}
\caption{General Framework}
\textbf{Input:} At time $t$ when decoding \\
\textbf{Output:} Input token $i_{t-1}$ 

\begin{algorithmic}[1]
\STATE{Sampling $p$ from $[0,1]$}
\IF{ $p<\epsilon_t$  }
	\STATE{$i_{t-1}=\hat{y}_{t-1}$}
\ELSE{ 
	\IF{ have not use $\hat{y}_{i}, i=1,2,...,t-2$ before}
		\STATE{$i_{t-1}=y_{t-1}$}
	\ELSE
		\STATE{Generating $y=\hat{y}^*_{t-1}$ by  {\it dynamic oracle}}
		\STATE{$i_{t-1}=\hat{y}^*_{t-1}$}
	\ENDIF
	}
\ENDIF
\end{algorithmic}
\end{algorithm}

\begin{table*}[h!]
\centering
\begin{tabular}{l|llllll}
\hline 
& MT02 &  MT03 & MT04  & MT05 & AVG & \\
\hline
 baseline & $33.55$ & $31.01$  & $33.28$ & $30.29$  & $32.03$ \\
 SS  & $33.47$ & $31.88$& $32.98$  & $30.72$  & $32.26$ \\
 SS+LM &  $34.56$ & $32.16$  & $33.36$ & $30.79$ & $32.71$\\
 SS+PRE &  $\textbf{34.63}$ & $\textbf{32.18}$  & $\textbf{34.18}$ & $\textbf{31.37}$ & $\textbf{33.09}$\\
\hline
\end{tabular}
\caption{\label{tbl:3}BLEU results}
\end{table*}

\subsection{ Language Model-based Dynamic Oracle}
Our first method utilize language model to guide the training procedure.

The goal of language modeling is to predict the next word in textual data
given specific context. Here we use recurrent neural
network based language models. 
Traditionally, backing-off language models
rely on the $n$-gram approximation,
which are often criticized because they could only store limited information and thus lack any explicit representation of long range dependency.
In contrast, 
Recurrent neural network language models always estimate probabilities based on the full history \cite{sundermeyer2012lstm}
In other words, recurrent neural networks
do not use limited size of context, which is the major
reason why we choose recurrent neural network language models.

Our first oracle is based on language model.
Basically, the well-trained language model will help us to generate ${y}^*_{t-1}$ given partially decoded sentence. Therefore, it could guarantee the coherence of our generated sentence. 

It should be noticed that in our scenario, even though the language model would predict
a probability distribution over the whole dictionary, we only select the 
word with the highest probability within the reference of the source sentence. In this way, we hope that the generated sentence
can be both coherent and precise.

\subsection{ Pre-trained Model-based Dynamic Oracle}
Our second method utilize a pre-trained neural machine translation model to guide the training procedure.

The basic procedure is similar to our first method
{\it i.e.} 
we randomly decide whether
to use the true previous token or the estimated token coming from the model itself. 
Once we choose an estimated word,
the next time when we decide to choose the true previous token, we provide the model with the token generated by the pre-trained model.

At this time, we do not have to select the word with the highest probability within the reference of the source sentence rather than the whole dictionary. Compared
to the previous method, this pre-trained model guided scheduled sampling not only utilize the information in the target sentence, but also utilize
the information in the source sentence. Therefore, we would expect the performance of the second method would be better than the first method.

Also, since at this time we do not limit our selections within the reference if source sentences, this model could add more diversity into the model and thus allow one sentence to have more translations.

\section{Experiment}

In this section we first describe the dataset used in our experiments, the training and evaluation details
, the baseline model we compare in experiments. And then we present the quantitive results of our experiments. At last, we would show some qualitative characteristics and demonstrate the potential of our model 

\subsection{Dataset and Setup}
We carry out experiments on a Chinese-English translation task. Our training data
for the translation task consists of 1.25M sentence pairs extracted from LDC corpora \footnote{The corpora include LDC2002E18, LDC2003E07, LDC2003E14, Hansards portion of LDC2004T07, LDC2004T08 and LDC2005T06.}, with 27.9M Chinese words and 34.5M English words respectively.
For our development set, we choose NIST 2002 dataset. Also, we choose the NIST 2003,
2004, 2005 as our test sets. 

Each neural machine translation model is trained using stochastic
gradient descent algorithm AdaGrad \cite{fazayeli2014adaptive}. We use mini-batch size of 32.
The word embedding dimension of source and target language is 600 and the size of hidden
layer is set to 1000. Also, for efficient training of the neural networks, we limit the source and target vocabularies
to the most frequent 30K words in Chinese and English
, covering approximately 97.7\% and
99.3\% of the two corpora respectively. 
All the out-of-vocabulary words are mapped to a special
token ``UNK".
We use case-insensitive 4-gram BLEU score as the evaluation metric \cite{papineni2002bleu}.

\subsection{Baseline Model}
In this work, our baseline neural machine translation system is
attentional encoder-decoder networks as implemented in DL4MT\footnote{https://github.com/nyu-dl/dl4mt-tutorial}, which is
an open source phrase-based translation system
available in github. 
In this framework, the baseline uses conditional
Gated Recurrent Unit (cGRU) with attention mechanism as the hidden unit type. The specific details of cGRU can also be found
in github \footnote{https://github.com/nyu-dl/dl4mt-tutorial/blob/master/docs/cgru.pdf}.

\subsection{Results}
\textbf{Table 1} shows the BLEU results for the baseline systems and
the dynamic oracle guided machine translation systems.
As we can see from the table, our method generally achieve better
performance than the baseline models. Moreover, the best average BLEU results improve more than
1 point on our dataset. Since the language model only utilizes the information in the target sentence, as we would expect, the pre-trained 
model behaves better. However, the language model-based oracle also improves the translation performance further testify the effectiveness of our algorithms.

\subsection{Case Studies}
In order to testify that our methods guide the system towards the right way, we sample a few examples, as illustrated in \textbf{Figure 2}.

As we can see, in the first sentence, we need to translate the Chinese sentence into
``the first round of border talks between ukraine and russia in moscow ." If we just use {\it Scheduled Sampling}, then at fourth time the 
model will make a mistake and generate``talks" instead of``border". Therefore, the next time if we still choose the original true previous token "talks", the sentence
will not be accurate. In such scenario, our {\it dynamic oracle} can be helpful. In this case, both the language model-based oracle and
pre-trained model-based oracle can generate  ``between" instead of  ``talks", which make the reference become accurate again. 

In the second sentence, we want our system to output ``esotina seeking nato membership .'', whereas
{\it Scheduled Sampling} generates ``esotina seeking to membership of". Here, the model makes a mistake by generating the word ``to". Again, as shown in the figure, our oracle can feed the model with 
the suitable word.

These examples demonstrate that our methods can be helpful in some cases and thus the improvement
in BLEU would be unsurprising.

\subsection{Analysis of Translation Results}
An important property of our second method, 
namely the pre-trained model-based dynamic oracle for NMT, 
is that it may provide the system with various 
 translations for each sentence. Traditionally, the model can only be trained with one unchangeable reference for each sentence while there may exist 
 a couple of correct references. In our second proposed method, since the baseline model may generate token which is not limited to single reference,
 we could add some diversity to NMT system.
 To verify our hypothesis, we present a few samples from the baseline model and our second method.

\begin{quote}
 \textbf{Reference}: japan temporarily freeze humanitarian assistance to russia .
 
\textbf{Baseline}: japan freezes its offer of humanitarian assistance to russia in the interim .
 
 \textbf{SS+PRE}: japan freezes humanitarian aid to russia for the time being .
 \end{quote}
 
As shown above, in the first example our reference is ``japan temporarily freeze humanitarian assistance to russia .''
Clearly, if we train
the model with our second method, it will output 
which is both precise and not
limited to the original reference.

Let us consider another example:
\begin{quote}
 \textbf{Reference}: at this time , the police have blocked the bombing scene .
 
\textbf{Baseline}: at this time , the police have UNK the blast at the scene .
 
 \textbf{SS+PRE}: the police have blocked the scene at the moment .
 \end{quote}
This example is another proof which shows that our second method could indeed add some diversity into NMT system.

\section{Conclusion and Future Work}
In this paper we highlight a major issue for NMT, {\it i.e.} the mismatch between where a model may end up in training and testing in the search space.

 {\it Scheduled Sampling} has been proposed to deal with this and we propose two methods to fix the drawbacks of the {\it Scheduled Sampling} . {\it Scheduled Sampling}  only feeds the previous predicted word into RNN decoder but still using the gold word for training, which is a static oracle fashion. The training oracle of our proposed method, on the other hand, is dynamic, according to a language model or a pre-trained NMT model. 

Also, as has been shown in Section $4.5$, our second model, namely the pre-trained model-based dynamic oracle for NMT, could provide the system with various 
 translations for each sentence. This feature is worth further research and could help enhance the performance of the model further. 
 
{\it Scheduled Sampling}  has been used in a variety of NLP tasks and it is curious that it does not help here. Although we have pointed out one potential problem, further investigation may help explain why this is the case.
\bibliographystyle{lrec}
\bibliography{xample}
\bibliographystylelanguageresource{lrec}
\bibliographylanguageresource{xample}

\end{document}